%% file: main.tex
\documentclass[letterpaper]{article} % DO NOT CHANGE THIS
\usepackage[preprint]{aaai2027}  % DO NOT CHANGE THIS
\usepackage[hyphens]{url}  % DO NOT CHANGE THIS
\usepackage{graphicx} % DO NOT CHANGE THIS
\usepackage{natbib}  % DO NOT CHANGE THIS AND DO NOT ADD ANY OPTIONS TO IT
\usepackage{caption} % DO NOT CHANGE THIS AND DO NOT ADD ANY OPTIONS TO IT
\usepackage{algorithm}
\usepackage{algorithmic}

\usepackage{newfloat}
\usepackage{listings}
\DeclareCaptionStyle{ruled}{labelfont=normalfont,labelsep=colon,strut=off} % DO NOT CHANGE THIS
\floatstyle{ruled}
\newfloat{listing}{tb}{lst}{}
\floatname{listing}{Listing}

\usepackage{booktabs}

\usepackage{multirow}
\usepackage{array}
\usepackage{makecell}

\usepackage{amsmath}
\usepackage{amssymb}
\usepackage{bm}

\title{ZoomDiff: A High-Fidelity Diffusion Model for Dual-Camera Smooth Zooming}
\author{
Jiayi Zhang\textsuperscript{\rm 1},
Renlong Wu\textsuperscript{\rm 1},
Yukang Ding\textsuperscript{\rm 2},
Sibin Deng\textsuperscript{\rm 2},
Wangmeng Zuo\textsuperscript{\rm 1}
}

\affiliations{
\textsuperscript{\rm 1}Harbin Institute of Technology\\
\textsuperscript{\rm 2}Taobao, Alibaba Group
}

\makeatletter
\DeclareRobustCommand\onedot{\futurelet\@let@token\@onedot}
\def\@onedot{\ifx\@let@token.\else.\null\fi\xspace}
 
\def\ie{\emph{i.e}\onedot}

\makeatother

\usepackage{xcolor}

\usepackage{booktabs, multirow, multicol, makecell}
\usepackage{graphicx}
\usepackage{amssymb}
\usepackage{xcolor}
\usepackage[abs]{overpic}
\usepackage{amsmath}
\RequirePackage{xspace}
\usepackage{cleveref} % 添加这行！

\usepackage{subcaption} % 用于创建子表
\usepackage{array}      % 扩展表格功能

\usepackage{booktabs}
\usepackage{multirow}
\usepackage{graphicx}
\usepackage{xcolor}
\usepackage{pifont}

\newcommand{\cmark}{\textcolor{green!60!black}{\ding{51}}}
\newcommand{\xmark}{\textcolor{red}{\ding{55}}}

\begin{document}

\maketitle

\begin{abstract}
Digital zoom transitions between dual cameras often exhibit conspicuous discontinuities in geometric structure and chromatic consistency, degrading the user experience.
While recent dual-camera smooth zoom (DCSZ) methods attempt to mitigate this by fine-tuning frame interpolation (FI) models on DCSZ data, they struggle with the large cross-view disparities and complex geometric transformations.
Considering that the generative prior of diffusion models is suitable for addressing this problem, we explore their application to DCSZ.
However, naively applying existing diffusion-based FI models still yields low-fidelity transitions due to insufficient conditional guidance, high-frequency information loss during VAE encoding, as well as inadequate temporal consistency.
To address this, we propose ZoomDiff, a high-fidelity diffusion model that leverages dual-camera inputs in both latent and pixel spaces for photo-realistic transitions.
Specifically, we first strengthen dual-image conditional guidance during the multi-step denoising process to improve geometric consistency.
Then we inject flow-aligned multi-scale features from the VAE encoder into the VAE decoder to recover high-frequency details, where flow-guided temporal consistency supervision are introduced to produce more smooth transitions.
Extensive experiments on both synthetic and real-world datasets demonstrate that ZoomDiff outperforms state-of-the-art methods quantitatively and qualitatively.
Project page: \url{https://jiayi-hit.github.io/ZoomDiff.github.io/}.

\end{abstract}

\section{Introduction}

\begin{figure}[t!]    
\centering    
\includegraphics[width=\linewidth]{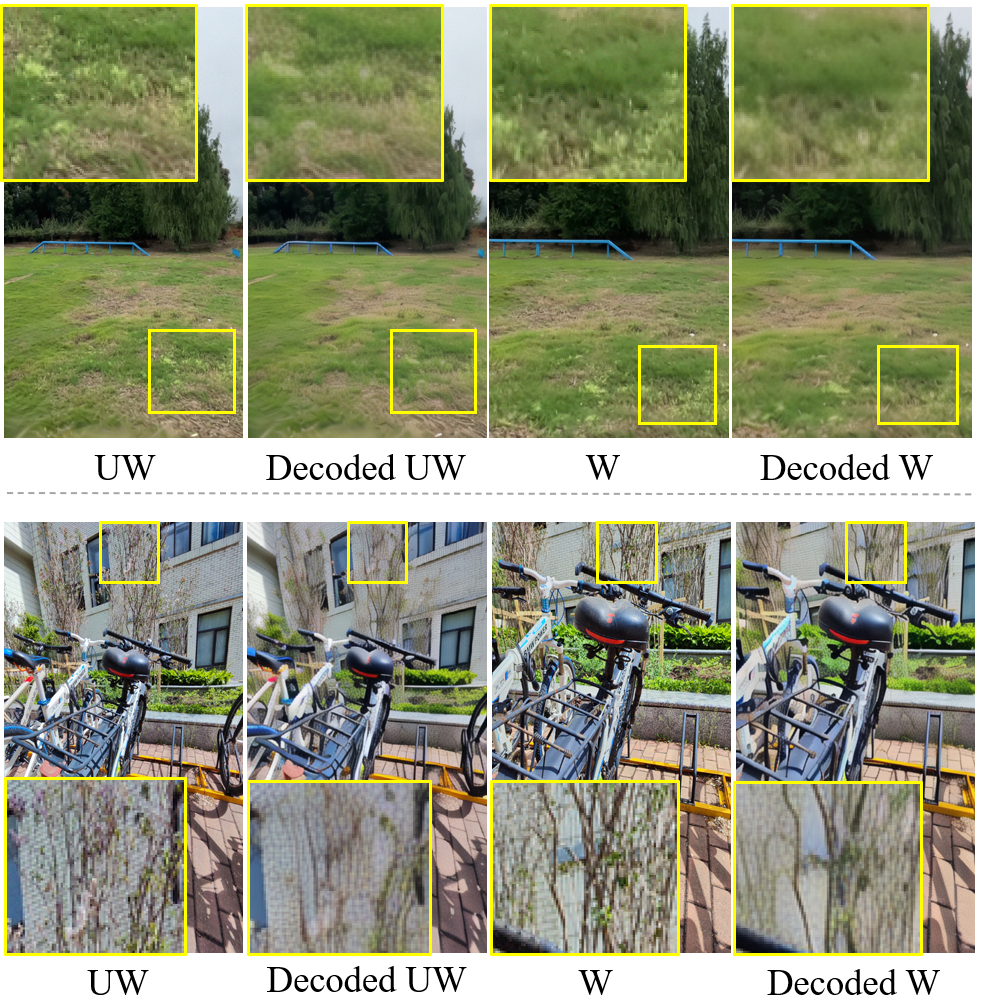}  
% \vspace{-6mm}
\caption{    
Illustration of high-frequency loss in the VAE latent space.
Images reconstructed from VAE latents contain less details than the original ones as indicated by yellows boxes.
} 
\label{fig:decode}
\end{figure}
Modern phones are equipped with multiple fixed-focal-length cameras for digital zoom.
Taking the dual-camera system as an example, it consists of a wide-angle (W) lens for primary imaging and an ultra-wide-angle (UW) lens for an even wider view.
Due to the distinct deployment locations and internal characteristics~\cite{zhang2022self,lee2022reference,alzayer2023dc2,abdelhamed2021leveraging} of the two cameras, a notable preview jump in geometric content and image color occurs when zooming between them, which significantly affects the user experience as they generally prefer a fluid preview.

To explore the issue, ~\cite{wu2024dual} first formulated that task as dual-camera smooth zoom (DCSZ), and tackled it from a data-centric perspective. 
They proposed a data construction pipeline based on 3DGS~\cite{kerbl20233d}, which generates synthetic DCSZ data to fine-tune a frame interpolation (FI) model for smooth transitions.
However, the results generally suffer from noticeable visual artifacts, which severely undermine the perceptual quality and zooming consistency.
We argue that the main cause lies in the inherent model-task mismatch.
Specifically, mainstream optical-flow-based~\cite{AMT, EDSC, IFRNet, RIFE, UPRNet} or diffusion-based~\cite{FCVG, framer, GI, TRF, wan} FI models are designed for videos with smaller camera or object motion, which cannot adapt to the large cross-view disparity and complex geometric transformations between dual cameras. 
In other words, a task-specific FI model tailored for the DCSZ task should be elaborately designed.

Compared to optical-flow-based~\cite{AMT, EDSC, IFRNet, RIFE, UPRNet} interpolation manners, diffusion-based ones~\cite{FCVG, framer, GI, TRF, wan} are more suited for such large cross-view transitions, due to the powerful generative prior.
However, naively applying them still encounters challenges in achieving high-fidelity transitions.
On the one hand, the insufficient utilization of dual-camera images as conditional guidance in the diffusion process leads to obvious spatial artifacts and color shift in the synthesized intermediate frames.
On the other hand, note that the VAE encoding process~\cite{rombach2022high} inevitably incurs high-frequency information loss.
As illustrated in Fig.~\ref{fig:decode}, images reconstructed from the VAE latent space contain less fine-grained details compared to the original ones.
Thus, only utilizing dual images in the latent space is insufficient to achieve high-fidelity zooming.

In this work, we propose ZoomDiff, a high-fidelity diffusion model for achieving smooth zoom between dual cameras, where the dual-camera images are fully employed in both latent and image spaces to address the aforementioned limitations.
On the one hand, in the latent space, the dual-camera latents are directly concatenated with the input noise on the endpoints and fed into the denoising network, while their semantic features are incorporated via cross-attention mechanisms. 
Additionally, to prevent error accumulation between the predicted dual-camera latents and the original ones during multi-step denoising, we explicitly replace the predicted dual-camera latents with the original ones at each denoising step.
% 
% concatenate the original latents with the predicted ones of intermediate frames .
% 
On the other hand, in the image space, we note that the multi-scale feature maps extracted by the VAE encoder preserve more high-frequency details, and their appropriate utilization is important for improving fidelity.
As the VAE encoding features of intermediate frames are unavailable, we first employ a pre-trained optical flow estimation network (e.g., SEA-RAFT~\cite{wang2024sea}) to estimate the motion between dual-camera images and the intermediate frames.
Then we warp the dual-image features to obtain features for the intermediate frames, which are injected into the VAE decoder via a ref injector module for better generation.
Besides, flow-guided temporal consistency supervision is adopted to produce more smooth transitions.

We conduct extensive experiments on both synthetic and real-world datasets.
The results demonstrate that our ZoomDiff achieves photo-realistic smooth zooming between dual cameras, outperforming state-of-the-art methods both quantitatively and qualitatively.
The contributions can be summarized as follows:
\begin{itemize}    
\item  We propose ZoomDiff, a high-fidelity diffusion model for achieving smooth zoom transitions between dual cameras.    
\item We suggest injecting the dual-camera images in both latent and image spaces, where latent information from the dual images is used in multiple ways to guide the denoising process for geometric consistency, while flow-aligned multi-scale features from the VAE encoder are injected into the VAE decoder to preserve high-frequency details.
\item Extensive experiments on both synthetic and real-world datasets demonstrate the effectiveness of ZoomDiff, outperforming state-of-the-art methods.
\end{itemize}

\section{Related Work}
\subsection{Dual-Camera Smooth Zoom}
They proposed a data construction pipeline based on 3DGS~\cite{kerbl20233d} that generates synthetic zooming sequences between dual cameras to fine-tune video frame interpolation (VFI) models~\cite{ AMT, UPRNet, RIFE, IFRNet} for smooth transitions.
However, even with data-driven adaptation, the synthesized results often exhibit boundary artifacts and structural inconsistencies due to substantial geometric disparity in such extremely large cross-view transitions.
These observations highlight the difficulty of the DCSZ task, and it is necessary to design an FI model tailored to it.

\begin{figure*}[t!]
    \centering
\includegraphics[width=0.99\linewidth]{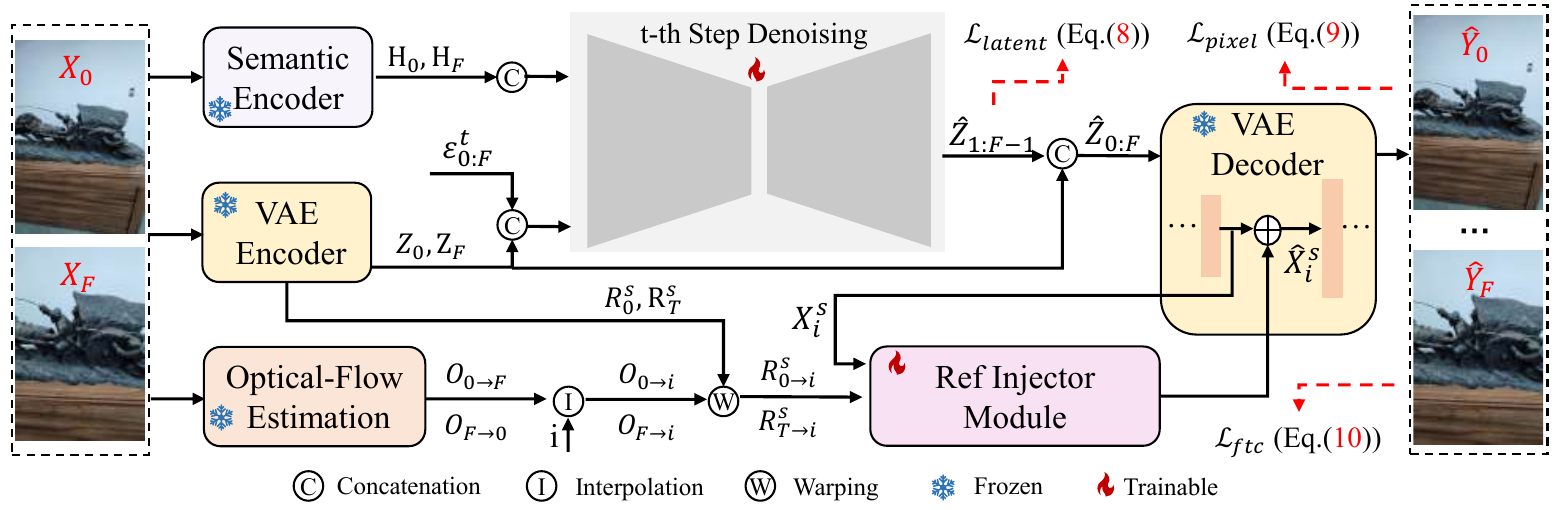}
    % \vspace{-3mm}
    \caption{
The dual-camera images (\ie, $\mathbf{X}_0$ and $\mathbf{X}_F$) are fed into the VAE encoder and the semantic encoder to extract latent representations (\ie, $\mathbf{Z}_0$ and $\mathbf{Z}_F$) and semantic representations (\ie, $\mathbf{H}_0$ and $\mathbf{H}_F$) respectively, which are incorporated into the denoising U-Net to guide the generation.
To avoid the high-frequency loss during VAE encoding, image features (\ie, $\mathbf{R}_{0}^{s}$ and $\mathbf{R}_{F}^{s}$) from the VAE encoder at $s$-th scale are injected into the VAE decoder.
Additionally, ZoomDiff is optimized by loss terms in both latent and pixel spaces.
    }
    \label{fig:method}
    % \vspace{-5mm}
\end{figure*}

\subsection{Optical-Flow-Based Frame Interpolation}
Driven by the robustness of optical flow, flow-based manners~\cite{RIFE, IFRNet,bao2019depth, bao2019memc, AMT, UPRNet, liu2017video,niklaus2018context, niklaus2020softmax, xu2019quadratic, EMAVFI, VFIFormer, BiFormer} currently dominate video frame interpolation, with extensive research into novel optimization objectives~\cite{RIFE, IFRNet} and advanced network designs~\cite{bao2019depth, bao2019memc, AMT, UPRNet, liu2017video, niklaus2018context, niklaus2020softmax, xu2019quadratic, EMAVFI, VFIFormer, BiFormer}.
To enhance flow accuracy, existing works have proposed various strategies, ranging from task-oriented flow distillation (e.g., RIFE~\cite{RIFE} and IFRNet~\cite{IFRNet}) and advanced bidirectional correlation retrieval (e.g., AMT~\cite{AMT} and UPRNet~\cite{UPRNet}) to unified motion-appearance extraction.
By relying on explicit motion modeling and geometric alignment, these methods excel at synthesizing temporally coherent frames under moderate motion. 
However, applying them to the DCSZ task reveals fundamental limitations. 
Dual-camera systems inherently introduce large cross-view disparities and progressive content occlusion. 
These characteristics severely violate the strict dense correspondence assumption fundamental to optical-flow estimation.
Consequently, traditional flow-based models often fail to produce reliable matches, struggling to maintain sharp boundaries and detailed structures in the DCSZ task.

\subsection{Diffusion-Based Frame Interpolation}
Recent works have extensively explored diffusion models~\cite{FCVG, framer, GI, TRF, vibid, wan} for generative frame interpolation without explicitly relying on optical flow.
Representative methods, such as FCVG~\cite{FCVG}, GI~\cite{GI}, and TRF~\cite{TRF}, formulate this as a conditional generation task by encoding the first and last frames as conditions for the denoising network. 
Furthermore, approaches such as VIBID~\cite{vibid}, Framer~\cite{framer}, and the recent video foundation model Wan2.1 (FLF)~\cite{wan} harness the rich generative priors of large-scale diffusion models, employing advanced temporal modules to synthesize coherent intermediate dynamics. 
By leveraging these generative capabilities, they can hallucinate plausible structures under large appearance changes, fundamentally alleviating the strict correspondence assumptions of flow-based methods.
However, naively applying existing diffusion-based FI models for the DCSZ task yields low-fidelity transitions, due to insufficient conditional guidance and high-frequency information loss during VAE encoding.
In this work, we propose ZoomDiff, a high-fidelity diffusion model that utilizes dual-camera inputs in both latent and pixel spaces for photo-realistic transitions. 

\section{Methods}

\subsection{Preliminaries}
\noindent \textbf{Problem Formation.}
Dual-Camera Smooth Zoom (DCSZ) aims to achieve a fluid zoom preview between dual-cameras on a mobile phone, which can be accomplished by a frame interpolation (FI) model to synthesize $F-1$ intermediate images $\{\hat{\mathbf{X}}_{i}\}_{i=1}^{F-1}$ between dual-camera images $\mathbf{X}_{0}$ and $\mathbf{X}_{F}$.
It can be written as,
\begin{equation}
\hat{\mathbf{{X}}}_{i} = \mathcal{FI}(\mathbf{X}_{0}, \mathbf{X}_{F}, i;\mathrm{\Theta_{\mathcal{FI}}}),
\end{equation}
where $\mathrm{\Theta_{\mathcal{FI}}}$ denotes the parameters of the FI model $\mathcal{FI}$. 
$i$ is the temporal index that indicates the relative relationship between the $i$-th synthesized image $\hat{\mathbf{X}}_{i}$ and the dual-camera images.
Compared to the widely explored video frame interpolation task that processes relatively smaller camera and object motion, DCSZ needs to handle the extremely large cross-view disparity and complex geometric transformations between dual cameras, making it more challenging.
\noindent \textbf{Diffusion Models.}
Generative diffusion models~\cite{blattmann2023stable,chen2023videocrafter1,wang2025lavie} iteratively denoise random Gaussian noise into high-quality images or videos by a denoising network. 
Taking Stable Video Diffusion (SVD)~\cite{blattmann2023stable}, it first encodes the given video using an autoencoder to obtain the latent representation $z$.
The forward process gradually adds noise to $z$ as follows,
\begin{equation}
    z_t = \alpha_t z + \sigma_t \epsilon, 
\end{equation}
where $\epsilon \sim \mathcal{N}(0, I)$, $\alpha_t$ and $\sigma_t$ represent the noise level
for denoising time $t$, defined by the noise schedule. 
For the backward process, a denoiser is used to iteratively denoise under the image-conditioning constraint. 
Finally, the generated videos are obtained through the VAE decoder.
Although several studies~\cite{FCVG, framer, GI, TRF, vibid} have endeavored to apply diffusion models to video frame interpolation, we observe that naively applying them to DCSZ still faces challenges in achieving high-fidelity transitions, due to insufficient conditional guidance for multi-step denoising and high-frequency information loss during VAE encoding. 
In this work, we propose ZoomDiff, a diffusion model based upon SVD~\cite{blattmann2023stable} for high-fidelity smooth zooming between dual cameras, as shown in Fig.~\ref{fig:method}.
Specifically, we first introduce several mechanisms to strengthen the conditional guidance of dual images during the multi-step denoising process in the latent space.
Then, we explore the dual-camera images in image space by incorporating multi-scale features from the VAE encoder into the VAE decoder to preserve high-frequency details
Finally, we provide the details of optimization objectives and inference acceleration.

\begin{table*}[!t]
\centering
\caption{Quantitative comparison results. $\uparrow$ denotes that higher values are better, while $\downarrow$ denotes that lower values are better. The best and second-best results are marked in \textbf{bold} and \underline{underlined}, respectively.}
\label{tab:main_results}

% \vspace{-1mm}
\small
\setlength{\tabcolsep}{3pt}
\renewcommand{\arraystretch}{1.02}

\resizebox{\textwidth}{!}{%
\begin{tabular}{m{1.6cm}m{2.0cm}ccccc@{\hspace{8pt}}cccc}
\toprule

\multirow{2}{*}{\centering Category}
&
\multirow{2}{*}{\centering Methods}
&
\multicolumn{5}{c}{Synthetic Dataset}
&
\multicolumn{4}{c}{Real-World Dataset}
\\

\cmidrule(lr){3-7}
\cmidrule(lr){8-11}

&
&
PSNR$\uparrow$
&
SSIM$\uparrow$
&
LPIPS$\downarrow$
&
PSNR-div$\uparrow$
&
FVD$\downarrow$
&
MUSIQ$\uparrow$
&
LIQE$\uparrow$
&
DBCNN$\uparrow$
&
DOVER$\uparrow$
\\

\midrule

\multirow{5}{*}{\makecell[c]{Optical-Flow\\Based}}
& RIFE
& 21.89 & 0.714 & \underline{0.261}
& \underline{20.51} & 352.214
& \underline{72.471} & \underline{4.816}
& \underline{0.681} & \underline{0.604}
\\

& EDSC
& 18.46 & 0.635 & 0.304
& 17.69 & 381.231
& 71.725 & 4.716 & 0.680 & 0.598
\\

& IFRNet
& 20.24 & 0.649 & 0.310
& 18.94 & 351.478
& 70.148 & 4.470 & 0.665 & 0.591
\\

& AMT
& 19.05 & 0.615 & 0.315
& 20.23 & 232.403
& 64.948 & 4.276 & 0.654 & 0.602
\\

& UPRNet
& \underline{22.58} & \underline{0.719} & 0.275
& 16.00 & 343.985
& 72.417 & 4.771 & 0.668 & 0.592
\\

\midrule

\multirow{7}{*}{\makecell[c]{Diffusion\\Based}}
& FCVG
& 20.20 & 0.633 & 0.343
& 19.65 & \underline{224.319}
& 68.325 & 4.544 & 0.537 & 0.519
\\

& GI
& 18.42 & 0.621 & 0.355
& 17.92 & 337.334
& 70.482 & 4.646 & 0.625 & 0.563
\\

& TRF
& 14.15 & 0.471 & 0.502
& 13.32 & 761.103
& 67.810 & 4.373 & 0.582 & 0.537
\\

& VIBID
& 17.90 & 0.579 & 0.428
& 16.96 & 585.608
& 71.347 & 4.552 & 0.657 & 0.556
\\

& Framer
& 19.95 & 0.634 & 0.363
& 18.85 & 239.095
& 53.837 & 2.431 & 0.391 & 0.335
\\

& Wan2.1
& 17.71 & 0.556 & 0.386
& 17.14 & 475.389
& 69.983 & 4.578 & 0.629 & 0.487
\\

& Ours
& \textbf{23.80} & \textbf{0.761} & \textbf{0.253}
& \textbf{23.01} & \textbf{131.332}
& \textbf{73.634} & \textbf{4.874}
& \textbf{0.692} & \textbf{0.665}
\\

\bottomrule
\end{tabular}%
}
% \vspace{-2mm}
\end{table*}

\subsection{Utilizing Dual Images in Latent Space}
\label{section:3_2}
On the one hand, we encode the dual images $\mathbf{X}_{0}$ and $\mathbf{X}_{F}$ using the pre-trained VAE encoder to obtain their latent representations $\mathbf{Z}_{0}$ and $\mathbf{Z}_{F}$. 
Then a condition sequence $\mathbf{Z}_{0:F}$ is concatenated with the Gaussian noise along the channel dimension and fed into the denoising U-Net, where $\mathbf{Z}_{0:F}$ can be written as,
\begin{equation}
\mathbf{Z}_{0:F} = [\, \mathbf{Z}_0,\; \mathbf{0},\; \ldots,\; \mathbf{0},\; \,\mathbf{Z}_F \,].
\end{equation}
This encourages the denoising process to follow a zoom trajectory anchored by dual images.
On the other hand, we feed $\mathbf{X}_{0}$ and $\mathbf{X}_{F}$ into a pre-trained CLIP~\cite{radford2021learningtransferablevisualmodels} encoder to extract semantic features $\mathbf{H}_{0}$ and $\mathbf{H}_{T}$. 
These features are then concatenated together and injected into the denoising U-Net via a cross-attention mechanism~\cite{lin2021catcrossattentionvision} to guide the generation process, which improves global visual coherence.
Additionally, we observe that errors between the predicted dual-camera latents and the original ones may accumulate during multi-step denoising, which easily leads to content and color shifts.
Considering the latents of dual images $\mathbf{Z}_{0}$ and $\mathbf{Z}_{F}$ are still available during inference, we directly replace the predicted latents of dual images $\hat{\mathbf{Z}}^{\,t}_{0}$ and $\hat{\mathbf{Z}}^{\,t}_{F}$ with $\mathbf{Z}_{0}$ and $\mathbf{Z}_{F}$ respectively at each $t$-th step of denoising, which can be written as, 
\begin{equation}
\hat{\mathbf{Z}}^{\,t}_{0} = \mathbf{Z}_0, \quad
\hat{\mathbf{Z}}^{\,t}_{F} = \mathbf{Z}_F.
\end{equation}
It explicitly anchors the latent diffusion process to dual images and stabilizes the generation of the intermediate frames.

\subsection{Utilizing Dual Images in Image Space}
Specifically, we deploy a pre-trained optical flow network (i.e., SEA-RAFT~\cite{wang2024sea}) to estimate the bidirectional optical flows (i.e., $\mathbf{O}_{0\rightarrow F}$ and $\mathbf{O}_{F\rightarrow 0}$) between the dual images. The optical flows at $i$-th intermediate timestamps (i.e., $\mathbf{O}_{0\rightarrow i}$ and $\mathbf{O}_{F\rightarrow i}$) are then obtained by linear interpolation.
It can be written as, 
\begin{equation}
O_{0\rightarrow i} =  \frac{i}{F}\, O_{0\rightarrow F}, \quad
O_{F\rightarrow i} = (1 - \frac{i}{F})\, O_{F\rightarrow 0}.
\end{equation}
After that, we back warp $\mathbf{R}_{0}^{s}$ and $\mathbf{R}_{F}^{s}$ with $\mathbf{O}_{0 \rightarrow i}$ and $\mathbf{O}_{F \rightarrow i}$ to obtain reference features of intermediate frames (i.e., $\mathbf{R}_{0 \rightarrow i}^s$ and $\mathbf{R}_{F \rightarrow i}^s$).
It can be written as,
\begin{equation}
\mathbf{R}_{0 \rightarrow i}^s = \mathcal{W}(R_0^s, O_{0 \rightarrow i}), \quad
\mathbf{R}_{F \rightarrow i}^s = \mathcal{W}(R_F^s, O_{F \rightarrow i}),
\end{equation}
where $\mathcal{W}(\cdot)$ denotes backward warping operation. 
Finally, $\mathbf{R}_{0 \rightarrow i}^s$, $\mathbf{R}_{F \rightarrow i}^s$ and the decoder features of $i$-th intermediate frame at $s$-th scale $\mathbf{D}_{i}^{s}$ are concatenated along channel dimension and then fed into a ref injector module to learn the high-frequency features, which are added with $\mathbf{D}_{i}^{s}$ to obtain the fused feature representations $\mathbf{\hat{D}}_{i}^{s}$.
It can be written as
\begin{equation}
\mathbf{\hat{D}}_{i}^{s} = \mathbf{D}_i^s + \mathcal{F}
\big(
\mathbf{R}_{0 \rightarrow i}^s,\;
\mathbf{R}_{F \rightarrow i}^s,\;
\mathbf{D}_i^s
\big),
\end{equation}
where $\mathcal{F}$ denotes the ref injector module.
It consists of four $3\times 3$ convolutional layers with ReLU activations followed by a 1×1 convolutional layer for channel reduction. 
$\mathbf{\hat{D}}_{i}^{s}$ is passed into the decoder at $s-1$ scale for latter reconstruction.

\subsection{Optimization Objectives}
\label{section:3_4}
We optimize ZoomDiff with the reconstruction loss in both latent and image spaces, $\mathcal{L}_{latent}$ and $\mathcal{L}_{pixel}$, respectively.
Besides, the flow-guided temporal consistency loss $\mathcal{L}_{ftc}$ is introduced to produce more smooth transitions.

$\mathcal{L}_{latent}$ measures the weighted reconstruction error between the predicted latent sequence $\hat{\mathbf{Z}}_{0:F}$ and the corresponding ground-truth sequence $\mathbf{Z}_{0:F}$, 
\begin{equation}
\mathcal{L}_{latent}
=
\sum_{i=0}^{F}
w_i
\left\|
\hat{\mathbf{Z}}_i-\mathbf{Z}_i
\right\|_2^2,
\end{equation}
where $w_i$ denotes the noise-dependent weighting coefficient following SVD~\cite{blattmann2023stable}.

$\mathcal{L}_{pixel}$ consists of a pixel-wise reconstruction loss and LPIPS-based perceptual loss, which can be written as,
\begin{equation}
\mathcal{L}_{pixel}
=
\sum_{i=0}^{F}
\left\|
\hat{\mathbf{X}}_i-\mathbf{X}_i
\right\|_2^2 + \lambda_{lpips}\sum_{i=0}^{F}
\phi
\left(
\hat{\mathbf{X}}_i,
\mathbf{X}_i
\right).
\end{equation}
$\phi$ denotes the LPIPS~\cite{zhang2018perceptual} network and $\lambda_{lpips}$ is set to 0.4.
$\mathcal{L}_{ftc}$ first employ the pretrained SEA-RAFT to estimate the optical flow $\mathbf{O}_{i+1\rightarrow i}$ from  $\mathbf{X}_{i+1}$ to $\mathbf{X}_i$, which is used to warp $\hat{\mathbf{X}}_i$ to $\hat{\mathbf{X}}_{i+1}$.
$\mathcal{L}_{\mathrm{ftc}}$ can be written as,
\begin{equation}
\mathcal{L}_{ftc}
=
\sum_{i=0}^{F-1}
\frac{
\left\|
\mathbf{M}_i
\odot
\left[
\mathcal{W}
\left(
\hat{\mathbf{X}}_i,
\mathbf{O}_{i+1\rightarrow i}
\right)
-
\hat{\mathbf{X}}_{i+1}
\right]
\right\|_1
}{
C\left\|\mathbf{M}_i\right\|_1+\epsilon
},
\end{equation}
where $\mathcal{W}$ denotes the backward warping operation, $\mathbf{M}_i$ is the valid warping mask, $C$ is the number of image channels, and $\epsilon$ is a small constant for numerical stability.

Overall, the loss used to optimize ZoomDiff is
\begin{equation}
\mathcal{L}_{zoom}
=
\mathcal{L}_{latent}
+
\lambda_{pixel}\mathcal{L}_{pixel} + \lambda_{ftc}\mathcal{L}_{ftc},
\end{equation}
where $\lambda_{pixel}$ and $\lambda_{ftc}$ are set to 1.0 and 0.05.
\subsection{Few-Step Adaptation}
To accelerate inference, we adapt the full-step model using the discrete noise levels of a target $K$-step scheduler.
Specifically, we define the scheduler-aligned noise set as,
\begin{equation}
\mathcal{S}_K
=
\left\{
\sigma_1,\sigma_2,\ldots,\sigma_K
\right\},
\end{equation}
where $\sigma_k$ denotes the noise level at the $k$-th sampling step. 
We uniformly sample $\sigma$ from $\mathcal{S}_K$ and construct the noisy latent sequence, which can be written as,
\begin{equation}
\sigma \sim \mathcal{U}(\mathcal{S}_K),
\qquad
\mathbf{Z}_{0:F}^{\sigma}
=
\mathbf{Z}_{0:F}
+
\sigma\boldsymbol{\epsilon},
\quad
\boldsymbol{\epsilon}
\sim
\mathcal{N}(\mathbf{0},\mathbf{I}).
\end{equation}
Following the EDM parameterization of SVD~\cite{blattmann2023stable}, the model predicts the corresponding clean latent sequence $\hat{\mathbf{Z}}_{0:F}$ and is optimized as,
\begin{equation}
\mathcal{L}_{\mathrm{latent}}^{K}
=
\mathbb{E}_{\mathbf{Z},\boldsymbol{\epsilon},
\sigma\sim\mathcal{U}(\mathcal{S}_K)}
\left[
w(\sigma)
\left\|
\hat{\mathbf{Z}}_{0:F}
-
\mathbf{Z}_{0:F}
\right\|_2^2
\right],
\end{equation}
where $w(\sigma)$ is the noise-dependent weighting function used in SVD~\cite{blattmann2023stable}. Restricting adaptation to the noise levels encountered by the target sampler improves generation stability under few sampling schedules.

\begin{figure*}[t!]    
\centering    
\includegraphics[width=0.99\textwidth]{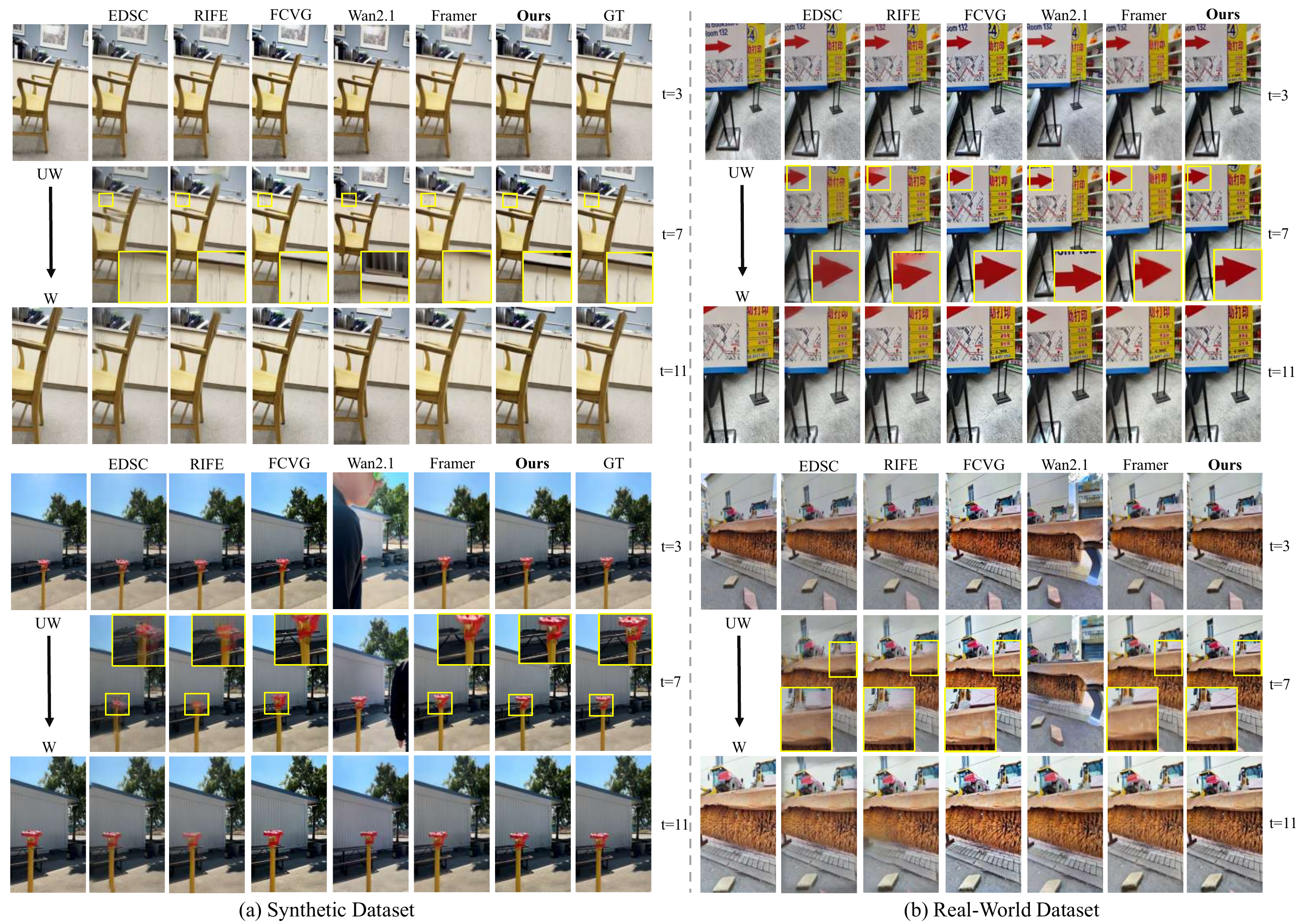}    
\caption{    
Visual comparisons on the synthetic dataset and the real-world dataset.
Our method still produces more consistent and visually pleasing transitions.
Please zoom in for better observation.
}    
\label{fig:qual}
\end{figure*}

\section{Experiments}
\subsection{Experimental Settings}
\noindent\textbf{Datasets.}

We follow ~\cite{wu2024dual} to construct 2100 sequences, including indoor and outdoor scenes, where 2000 sequences are used for training, and the non-overlapped 100 ones are used for evaluation.
Each sequence comprises 14 frames at a resolution of $576 \times 320$.
Besides, we use 100 real-world scenes captured by Redmi K50 Ultra from ~\cite{wu2024dual} to evaluate performance in the real world.
\noindent\textbf{Training Configurations.}
For stable optimization, we adopt a two-stage training strategy.
In the first stage, we train the denoising U-Net for 40k iterations for coarse geometry optimization.
In the second stage, the denoising U-Net is frozen, and only the parameters of the ref injector module are optimized for an additional 10k iterations for fine-grained details.
The learning rate is set to $3 \times 10^{-5}$ and the Adam optimizer~\cite{kingma2014adam} is adopted.
We augment the training data with random horizontal and vertical flips.
All experiments are conducted with PyTorch~\cite{paszke2019pytorch} on two Nvidia GeForce RTX A6000 GPUs.
\noindent\textbf{Evaluation Configurations.}
During inference, we set the random seed to $42$ for deterministic evaluation.
For the synthetic dataset with ground-truth intermediate frames, we calculate full-reference metrics, including PSNR~\cite{huynh2008scope}, SSIM~\cite{wang2004image}, and LPIPS~\cite{zhang2018unreasonable} for spatial fidelity, as well as PSNR-div~\cite{daly2025efficient} and FVD~\cite{unterthiner2018towards} for temporal consistency evaluation.
For real-world data without ground-truth references, we employ several no-reference image and video quality assessment metrics, including MUSIQ~\cite{ke2021musiq}, LIQE~\cite{zhang2023blind}, DBCNN~\cite{wang2023exploring}, and DOVER~\cite{wu2023exploring}.

\subsection{Comparison with State-of-the-Art Methods}

\noindent\textbf{Comparison Configurations}.
We compare our method with 12 state-of-the-art frame interpolation methods, including 5 optical-flow-based ones (i.e., EDSC~\cite{EDSC}, IFRNet~\cite{IFRNet}, RIFE~\cite{RIFE}, AMT~\cite{AMT}, UPRNet~\cite{UPRNet} and 6 diffusion-based ones (i.e., FCVG~\cite{FCVG}, GI~\cite{GI}, TRF~\cite{TRF}, VIBID~\cite{vibid}, Framer~\cite{framer} and Wan2.1~\cite{wan}).
Note that Wan2.1~\cite{wan} refers to the first-last-frame-to-video version (FLF2V-14B-720P).
\noindent\textbf{Quantitative Results}.
Table~\ref{tab:main_results} presents the quantitative comparison against state-of-the-art methods. 
On the synthetic dataset, ZoomDiff achieves the best reconstruction fidelity and temporal consistency, outperforming all competitors with PSNR of 23.80, SSIM of 0.761, LPIPS of 0.253, PSNR-div of 23.01, and FVD of 131.332.
Furthermore, on the real-world dataset, ZoomDiff exhibits superior generalization capability and visual quality. 
It establishes new state-of-the-art performance across all no-reference image quality assessment metrics. 
Overall, these results validate the superiority of ZoomDiff over existing methods for generating high-fidelity, temporally consistent, and visually pleasing results in both synthetic and complex real-world scenarios.

\noindent\textbf{Qualitative Results}.
The qualitative comparisons on synthetic and real-world datasets are presented in Fig.~\ref{fig:qual}.
Optical-flow-based methods~\cite{AMT,EDSC,IFRNet,RIFE,UPRNet} exhibit noticeable structural artifacts, degrading the zooming consistency.
Meanwhile, diffusion-based approaches~\cite{FCVG,framer,GI,wan} generate inconsistent geometry and confusing details due to inadequate dual-camera guidance and VAE-induced high-frequency loss.
In contrast, by jointly exploiting the dual-camera information in both latent and image spaces, ZoomDiff produces geometrically consistent zoom transitions with photo-realistic details and fewer visual artifacts.
Additional video results are provided in \textit{Suppl.}.

\section{Ablation Study}
We conduct ablation studies to analyze the effectiveness of each component in the proposed ZoomDiff on both synthetic and real-world datasets.
Specifically, we first investigate different strategies for injecting the latents of dual-image into the diffusion process.
Then, we evaluate the effectiveness of injecting multi-scale image features from the VAE encoder into the VAE decoder.
Finally, we study the impact of optimization objectives and few-step adaptation.

\subsection{Effect of Dual-Image Utilization in Latent Space}
\label{section:5_1}
We exploit dual-image information in the latent space through three complementary designs.
First, the encoded latents of the two reference images are concatenated with the noisy latents at the first and last frames, respectively.
Second, their semantic features are injected into the denoising network through cross-attention to provide high-level structural guidance.
Third, the predicted boundary latents are replaced with the original reference latents at each denoising step, preventing errors at the two endpoints from accumulating throughout the sequence.
We validate these designs through the ablations in Table~\ref{tab:ablate_failed_w}.
Removing the semantic features (\ie, w/o Semantic) weakens structural fidelity, indicating that latent inputs alone are insufficient to preserve high-level content.
Without step-wise latent replacement (\ie, w/o Replacement), prediction errors at the boundary frames gradually accumulate, leading to brightness deviations and temporally propagated artifacts.
We also investigate a naive strategy that concatenates both reference latents with every noisy latent along the channel dimension (\ie, Naive Concat).
This indiscriminate fusion disrupts the original denoising distribution and introduces inconsistent conditioning across frames, resulting in inferior reconstruction quality and unstable brightness.
By combining semantic guidance with explicit boundary anchoring, our full model achieves the best overall performance across the synthetic and real-world datasets.

\begin{table}[t!]
\centering
\caption{Effect of different strategies to inject dual images into the latent space.}
\label{tab:ablate_failed_w}

\small
\setlength{\tabcolsep}{2.2pt}
\renewcommand{\arraystretch}{1.05}

\resizebox{\columnwidth}{!}{%
\begin{tabular}{
@{}
l
@{\hspace{3pt}}
ccccc
@{\hspace{9pt}}
cccc
@{}
}
\toprule

% \multirow{2}{*}{\multicolumn{1}{c}{Methods}}
\multirow{2}{*}{Methods}
&
\multicolumn{5}{c}{Synthetic Dataset}
&
\multicolumn{4}{c}{Real-World Dataset}
\\

\cmidrule(lr){2-6}
\cmidrule(lr){7-10}

&
PSNR$\uparrow$
&
SSIM$\uparrow$
&
LPIPS$\downarrow$
&
PSNR-div$\uparrow$
&
FVD$\downarrow$
&
MUSIQ$\uparrow$
&
LIQE$\uparrow$
&
DBCNN$\uparrow$
&
DOVER$\uparrow$
\\

\midrule

w/o Semantic
& \underline{23.60}
& 0.755
& 0.261
& \underline{22.80}
& \underline{140.562}
& 73.287
& 4.814
& 0.671
& \underline{0.650}
\\

w/o Replacement
& 22.96
& \underline{0.758}
& \underline{0.255}
& 22.16
& 148.340
& \underline{73.423}
& \underline{4.870}
& \underline{0.686}
& \textbf{0.665}
\\

Naive Concat
&20.90
&0.702
&0.321
&19.98
&245.214
&71.393
&4.678
&0.630
&0.648
\\

\midrule

Ours
& \textbf{23.80}
& \textbf{0.761}
& \textbf{0.253}
& \textbf{23.01}
& \textbf{131.332}
& \textbf{73.634}
& \textbf{4.874}
& \textbf{0.692}
& \textbf{0.665}
\\

\bottomrule
\end{tabular}%
}
\end{table}

\subsection{Effect of Dual-Image Utilization in Image Space}
\label{section:5_2}
\begin{figure}[t!]    
\centering    
\includegraphics[width=\columnwidth]{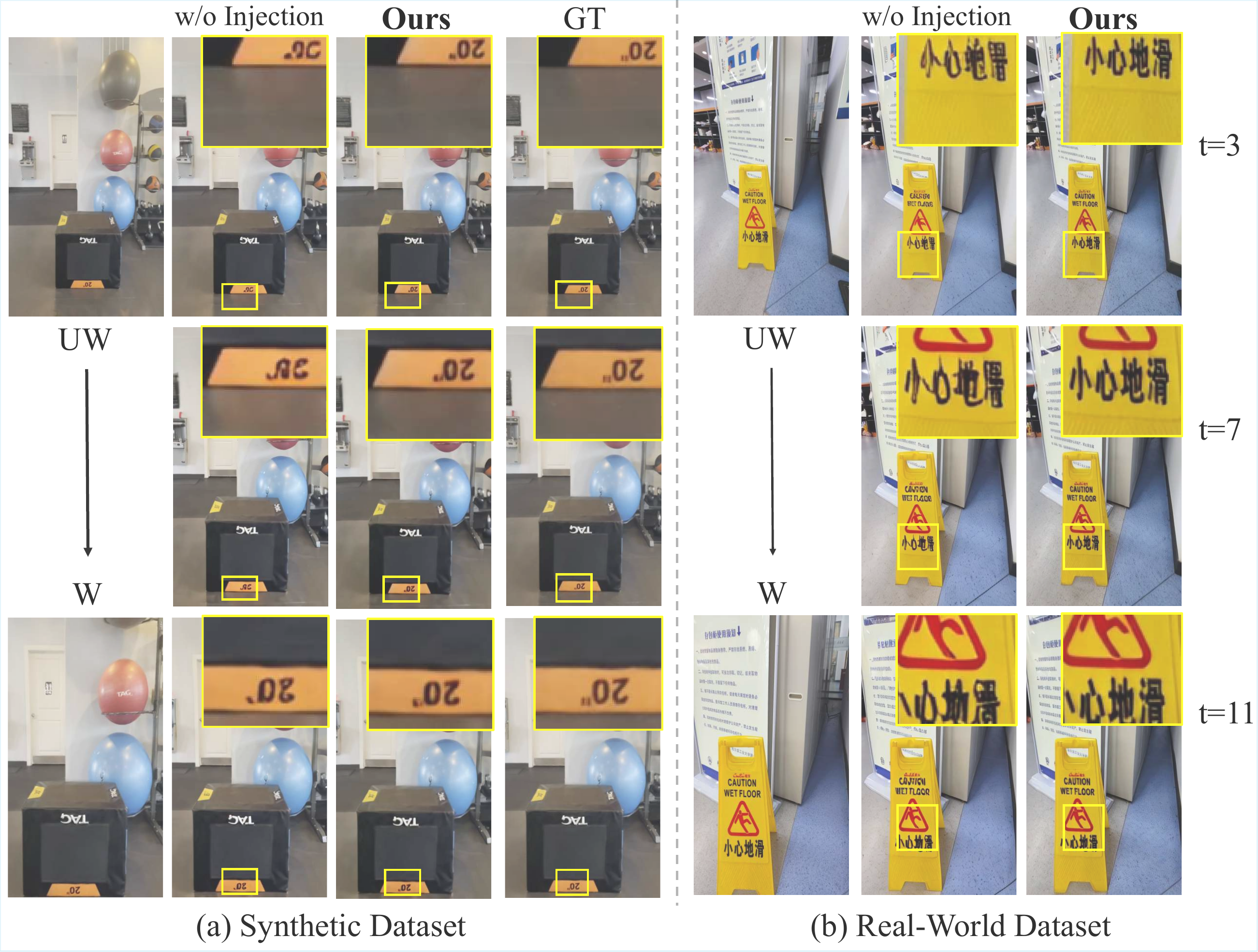}  
% \vspace{-6mm}
\caption{  
Visual effect of injecting multi-scale features of dual images from the VAE encoder into the VAE decoder.
Please zoom in for better observation.
} 
\label{fig:inject}

\end{figure}

To compensate for the high-frequency information lost during VAE encoding, we inject the multi-scale features of the dual-camera images from the VAE encoder into the corresponding layers of the VAE decoder.
We evaluate this design by removing the feature injection while keeping the remaining components unchanged.
The qualitative and quantitative comparisons on the synthetic and real-world datasets are presented in Fig.~\ref{fig:inject} and Table~\ref{tab:ablate_decoder}, respectively.
As shown in Fig.~\ref{fig:inject}, removing the encoder features (\ie, w/o Injection) leads to blurred textures, weakened fine structures, and local distortions, consistent with the information loss introduced by the VAE bottleneck in Fig.~\ref{fig:decode}.
In contrast, ZoomDiff provides the decoder with complementary multi-scale information from both input images, thereby preserving spatial structures and recovering sharper and more faithful details.
The consistent improvements across the quantitative metrics in Table~\ref{tab:ablate_decoder} further demonstrate the importance of utilizing dual-image information in the image space.

\begin{table}[t!]
\centering
\caption{Effect of injecting multi-scale features of dual images from the VAE encoder into the VAE decoder.}
\label{tab:ablate_decoder}

\small
\setlength{\tabcolsep}{2.2pt}
\renewcommand{\arraystretch}{1.05}

\resizebox{\columnwidth}{!}{%
\begin{tabular}{
@{}
l
@{\hspace{3pt}}
ccccc
@{\hspace{9pt}}
cccc
@{}
}
\toprule

\multicolumn{1}{c}{\multirow{2}{*}{Method}}
&
\multicolumn{5}{c}{Synthetic Dataset}
&
\multicolumn{4}{c}{Real-World Dataset}
\\

\cmidrule(lr){2-6}
\cmidrule(lr){7-10}

&
PSNR$\uparrow$
&
SSIM$\uparrow$
&
LPIPS$\downarrow$
&
PSNR-div$\uparrow$
&
FVD$\downarrow$
&
MUSIQ$\uparrow$
&
LIQE$\uparrow$
&
DBCNN$\uparrow$
&
DOVER$\uparrow$
\\

\midrule

w/o Injection
& 23.33
& 0.742
& 0.267
& 22.47
& 158.369
& 73.113
& 4.827
& 0.657
& 0.650
\\

Ours
& \textbf{23.80}
& \textbf{0.761}
& \textbf{0.253}
& \textbf{23.01}
& \textbf{131.332}
& \textbf{73.634}
& \textbf{4.874}
& \textbf{0.692}
& \textbf{0.665}
\\

\bottomrule
\end{tabular}%
}

% \vspace{-4mm}
\end{table}

\subsection{Effect of Optimization Objectives}
\label{section:5_3}
We introduce $\mathcal{L}_{latent}$ and $\mathcal{L}_{pixel}$ to optimize the proposed ZoomDiff in the latent space and image space respectively, where $\mathcal{L}_{pixel}$ is the combination of a pixel-wise reconstruction loss $\mathcal{L}_{mse}$, an LPIPS-based perceptual loss $\mathcal{L}_{lpips}$, and the flow-guided temporal consistency loss $\mathcal{L}_{ftc}$.
As shown in Table~\ref{tab:ablate_loss}, removing both $\mathcal{L}_{lpips}$ and $\mathcal{L}_{ftc}$ slightly improves PSNR and SSIM metrics, but degrades all perceptual metrics, demonstrating that pixel-wise supervision alone is insufficient for generating visually pleasing results.
Introducing $\mathcal{L}_{lpips}$  consistently improves perceptual quality, demonstrating the value of perceptual supervision for recovering fine-grained image details.
Further incorporating $\mathcal{L}_{ftc}$ achieves the best overall performance, validating the effectiveness of the proposed flow-guided temporal consistency supervision.
Visual comparisons in the \textit{Suppl} further demonstrate that $\mathcal{L}_{lpips}$ helps generate more fine-grained details, while $\mathcal{L}_{ftc}$ effectively suppresses temporal flickering.

\begin{table}[t]
\centering
\caption{Effect of different optimization objectives.}
\label{tab:ablate_loss}

\setlength{\tabcolsep}{1.8pt}
\renewcommand{\arraystretch}{1.08}

\resizebox{\columnwidth}{!}{%
\begin{tabular}{
@{}
cc
@{\hspace{4pt}}
ccccc
@{\hspace{5pt}}
cccc
@{}
}
\toprule

\multirow{2}{*}{$\mathcal{L}_{lpips}$}
&
\multirow{2}{*}{$\mathcal{L}_{ftc}$}
&
\multicolumn{5}{c}{Synthetic Dataset}
&
\multicolumn{4}{c}{Real-World Dataset}
\\

\cmidrule(lr){3-7}
\cmidrule(lr){8-11}

&
&
\shortstack{PSNR$\uparrow$}
&
\shortstack{SSIM$\uparrow$}
&
\shortstack{LPIPS$\downarrow$}
&
\shortstack{PSNR-div$\uparrow$}
&
\shortstack{FVD$\downarrow$}
&
\shortstack{MUSIQ$\uparrow$}
&
\shortstack{LIQE$\uparrow$}
&
\shortstack{DBCNN$\uparrow$}
&
\shortstack{DOVER$\uparrow$}
\\

\midrule

\xmark
&
\xmark
&
23.62
&
\textbf{0.762}
&
0.274
&
22.71
&
156.315
&
68.619
&
4.521
&
0.534
&
0.570
\\

\xmark
&
\cmark
&
\underline{23.69}
&
\underline{0.761}
&
0.283
&
\underline{22.74}
&
\underline{139.517}
&
69.150
&
4.583
&
0.536
&
0.555
\\

\cmark
&
\xmark
&
23.57
&
0.756
&
\textbf{0.247}
&
22.73
&
145.226
&
\underline{73.554}
&
\underline{4.845}
&
\underline{0.674}
&
\underline{0.647}
\\

\cmark
&
\cmark
&
\textbf{23.80}
&
\underline{0.761}
&
\underline{0.253}
&
\textbf{23.01}
&
\textbf{131.332}
&
\textbf{73.634}
&
\textbf{4.874}
&
\textbf{0.692}
&
\textbf{0.665}
\\

\bottomrule
\end{tabular}%
}

\end{table}

\subsection{Effect of Few-Step Adaptation}

The iterative denoising process of diffusion models incurs substantial inference overhead. We therefore adapt the converged 25-step model to 8-step and 4-step sampling schedules without modifying the network architecture. As shown in Table~\ref{tab:few_step_efficiency}, the resulting models achieve $2.16\times$ and $2.92\times$ inference speedups, respectively.
Table~\ref{tab:few_step} demonstrates the effectiveness of the proposed adaptation. Compared with directly shortening the sampling trajectory, the adapted 8-step and 4-step models consistently improve all evaluation metrics. In particular, the adapted 8-step model surpasses the original 25-step model on all synthetic metrics while maintaining comparable performance on the real-world dataset. Although the 4-step model shows a moderate quality degradation relative to the 25-step model, it remains clearly superior to the unadapted 4-step variant. These results demonstrate that few-step adaptation substantially improves inference efficiency while preserving generation quality.

\begin{table}[t]
\centering
\caption{Inference efficiency comparison of ZoomDiff.}
\label{tab:few_step_efficiency}

\small
\setlength{\tabcolsep}{4pt}
\renewcommand{\arraystretch}{0.75}

\resizebox{\columnwidth}{!}{%
\begin{tabular}{lccccc}
\toprule
Steps & \#Params (M) & \#FLOPs (T) & Runtime (s)$\downarrow$ & Speedup$\uparrow$\\
\midrule
25 & 2091 & 743 & 45.89 & 1.00$\times$\\
 8  & 2091 & 381 & 21.28 & 2.16$\times$\\
4  & 2091 & 296 & 15.71 & 2.92$\times$\\
\bottomrule
\end{tabular}
}
\end{table}

\begin{table}[t!]
\centering
\caption{Effect of few-step adaptation under different sampling steps.}
\label{tab:few_step}

\small
\setlength{\tabcolsep}{2pt}
\renewcommand{\arraystretch}{1.05}

\resizebox{\columnwidth}{!}{%
\begin{tabular}{
@{}
l
@{\hspace{3pt}}
ccccc
@{\hspace{8pt}}
cccc
@{}
}
\toprule

\multirow{2}{*}{\centering Method}
&
\multicolumn{5}{c}{Synthetic Dataset}
&
\multicolumn{4}{c}{Real-World Dataset}
\\

\cmidrule(lr){2-6}
\cmidrule(lr){7-10}

&
PSNR$\uparrow$
&
SSIM$\uparrow$
&
LPIPS$\downarrow$
&
PSNR-div$\uparrow$
&
FVD$\downarrow$
&
MUSIQ$\uparrow$
&
LIQE$\uparrow$
&
DBCNN$\uparrow$
&
DOVER$\uparrow$
\\

\midrule

Ours (25-step)
& 23.80
& 0.761
& 0.253
& 23.01
& \underline{131.332}
& \textbf{73.634}
& \textbf{4.874}
& \textbf{0.692}
& \textbf{0.665}
\\
\midrule

w/o Adaptation (8-step)
& 22.32
& 0.732
& 0.257
& 22.63
& 140.904
& 72.463
& 4.782
& 0.673
& 0.640

\\

Ours (8-step)
& \textbf{24.08}
& \textbf{0.787}
& \textbf{0.228}
& \textbf{24.45}
& \textbf{118.246}
& 72.830
& \underline{4.806}
& \underline{0.689}
& 0.646
\\
\midrule

w/o Adaptation (4-step)
& 22.37
& 0.712
& 0.266
& 22.16
& 218.346
& 72.423
& 4.789
& 0.667
& 0.641
\\

Ours (4-step)
& \underline{23.90}
& \underline{0.766}
& \underline{0.242}
& \underline{23.25}
& 184.081
& \underline{72.990}
& \underline{4.806}
& 0.675
& \underline{0.647}
\\

\bottomrule
\end{tabular}%
}
\end{table}

\section{Conclusion}
In this work, we present ZoomDiff, a high-fidelity diffusion framework tailored for the Dual-Camera Smooth Zoom (DCSZ) task, where the dual-camera images are fully utilized in both the latent and the image space to improve fidelity.
Specifically, we introduce multiple ways to utilize the latent information from the dual images to guide the denoising process for geometric consistency, and inject flow-aligned multi-scale features from the VAE encoder into the VAE decoder via a ref injector module to preserve high-frequency details.
Extensive experiments on both synthetic and real-world datasets demonstrate that ZoomDiff outperforms state-of-the-art methods, achieving photo-realistic smooth zooming transitions between dual cameras.

% =========================
% Supplementary Material
% =========================

\clearpage

% \begin{center}
% {\LARGE\bfseries Supplementary Material}\\[0.5em]
% {\large\bfseries
% ZoomDiff: A High-Fidelity Diffusion Model for Dual-Camera Smooth Zooming}
% \end{center}

\twocolumn[{
  \centering
  {\LARGE\bfseries Supplementary Material\par}
  \vskip 0.5em
  {\large\bfseries
  ZoomDiff: A High-Fidelity Diffusion Model for Dual-Camera Smooth Zooming\par}
  \vskip 1em
}]

\vspace{1em}

\input{supp}

\bibliography{aaai2027}

\end{document}

%% file: supp.tex
\noindent{The content of the supplementary material involves:}
\vspace{1mm}
\begin{itemize}
\item More Comparisons in Sec.A.
\vspace{1mm}
\item More ablation results in Sec.B.
\vspace{1mm}
\item More challenging cases in Sec.C.
% \vspace{1mm}
% \item More challenging cases in Sec.D.
\end{itemize}

% \section{Sec.A. Video Demonstrations}
\section{Sec.A More Comparisons}
Supplementary video results are provided in the submitted multimedia materials (HTML-based visualization page).
The visualization page presents qualitative results of ZoomDiff on synthetic and real-world datasets, including cases with different levels of scene complexity.
% 
% \section{Sec.B. Qualitative Comparisons}
% \section{Sec.B}

\begin{figure*}[t!]
\centering
\includegraphics[width=0.99\textwidth]{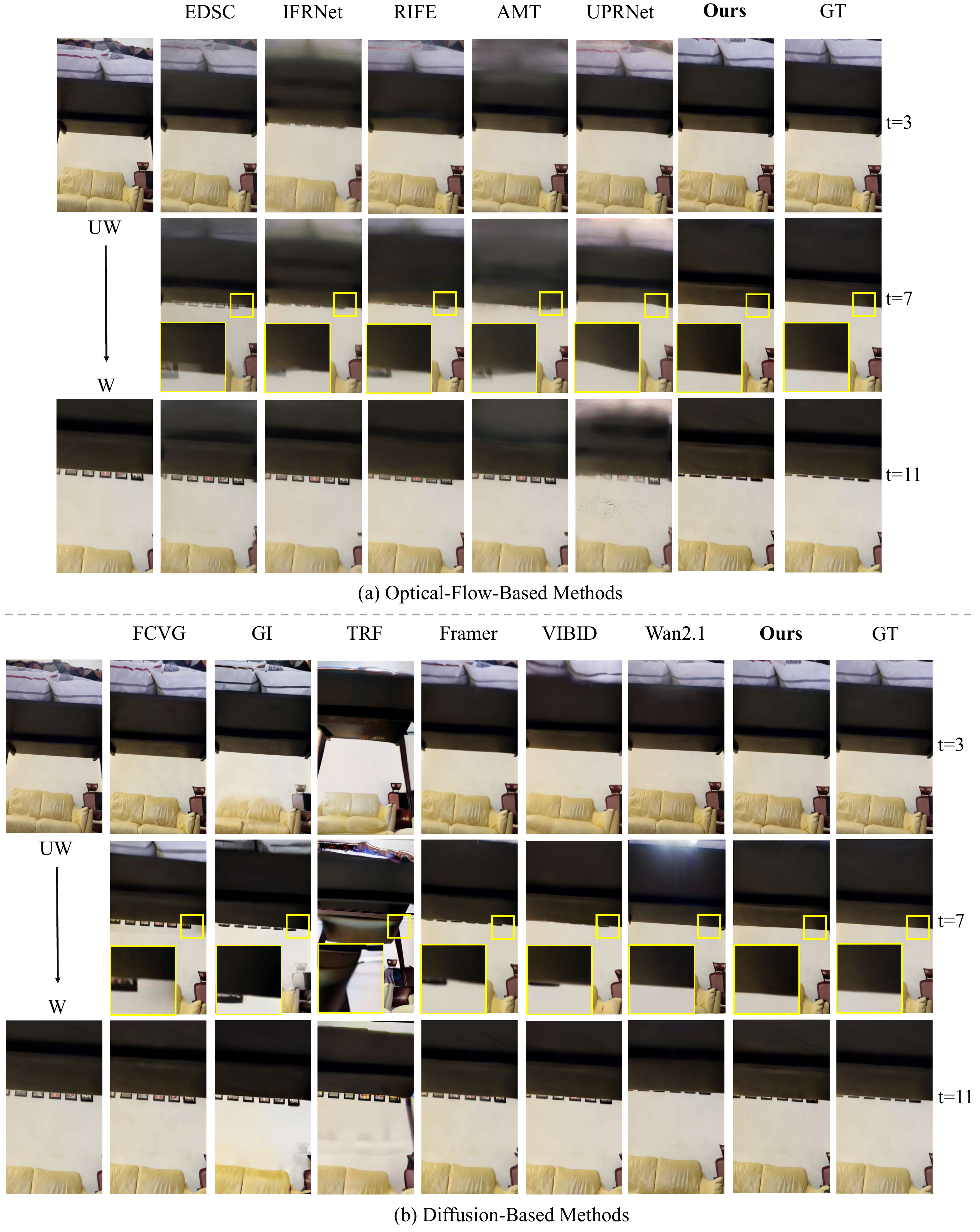}
\caption{
Visual comparison with state-of-the-art frame interpolation methods on a synthetic scene.
(a) Optical-flow-based methods.
(b) Diffusion-based methods.
Our ZoomDiff generates more geometrically consistent zoom transitions with fewer artifacts (yellow boxes).
Please zoom in for better observation.
}
\label{fig:compare_syn1}
\end{figure*}

\begin{figure*}[t!]
\centering
\includegraphics[width=0.99\textwidth]{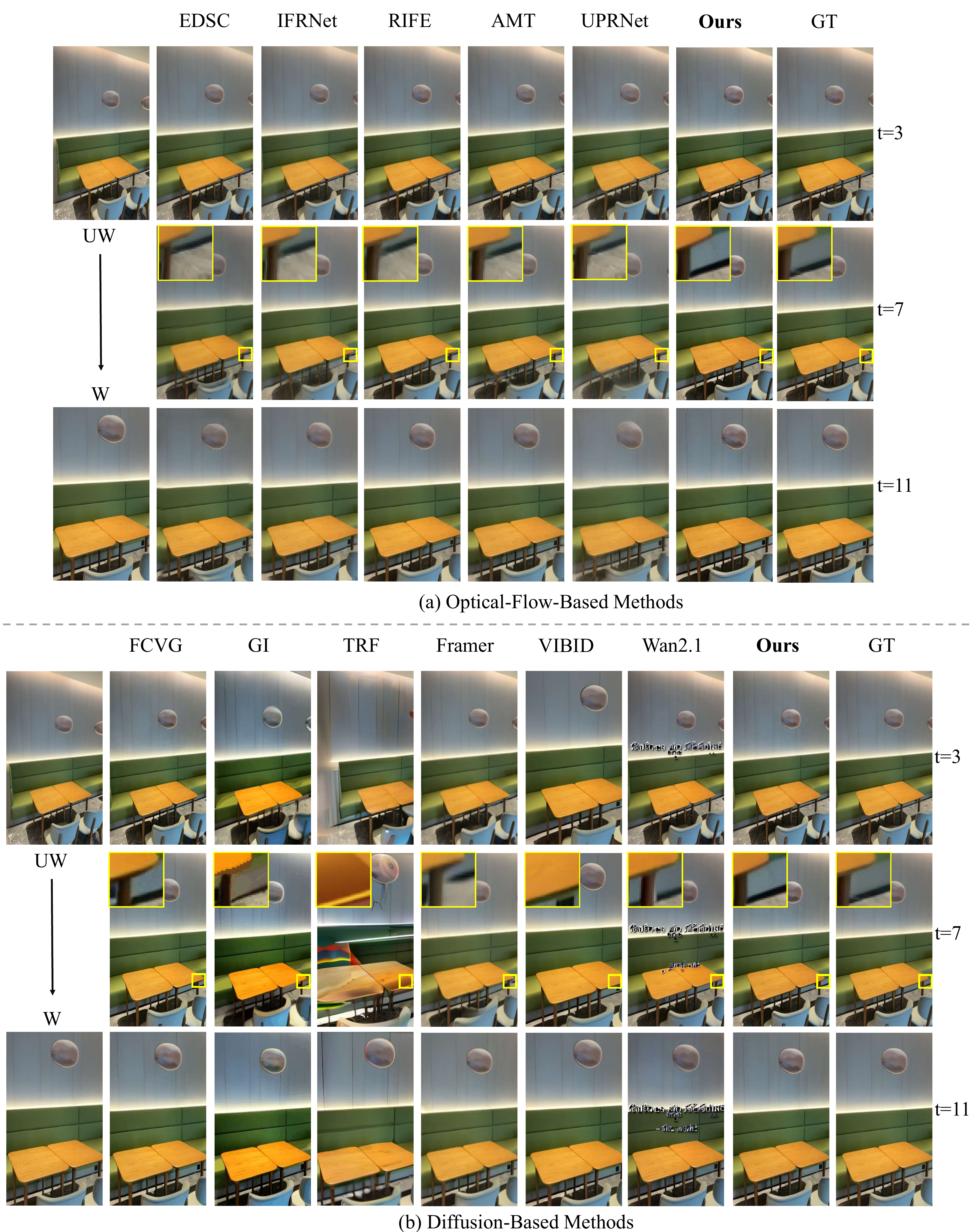}
\caption{
Visual comparison with state-of-the-art frame interpolation methods on a synthetic scene.
(a) Optical-flow-based methods.
(b) Diffusion-based methods.
Our ZoomDiff generates more geometrically consistent zoom transitions with fewer artifacts (yellow boxes).
Please zoom in for better observation.
}
\label{fig:compare_syn2}
\end{figure*}

\begin{figure*}[t!]
\centering
\includegraphics[width=0.90\textwidth]{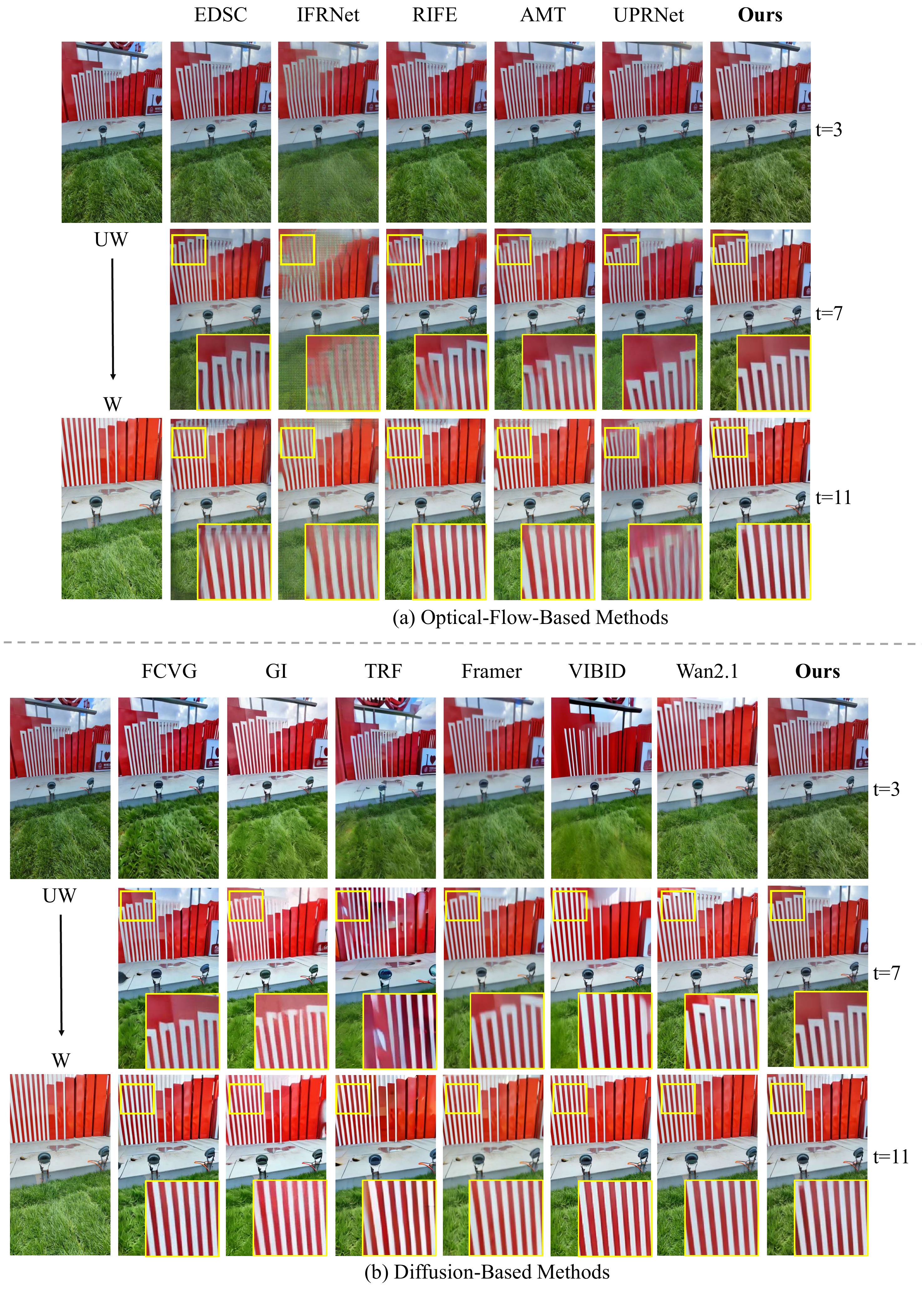}
\caption{
Visual comparison with state-of-the-art frame interpolation methods on a real-world scene.
(a) Optical-flow-based methods.
(b) Diffusion-based methods.
Our ZoomDiff generates more geometrically consistent zoom transitions with fewer artifacts (yellow boxes).
Please zoom in for better observation.
}
\label{fig:compare_real1}
\end{figure*}

\begin{figure*}[t!]
\centering
\includegraphics[width=0.90\textwidth]{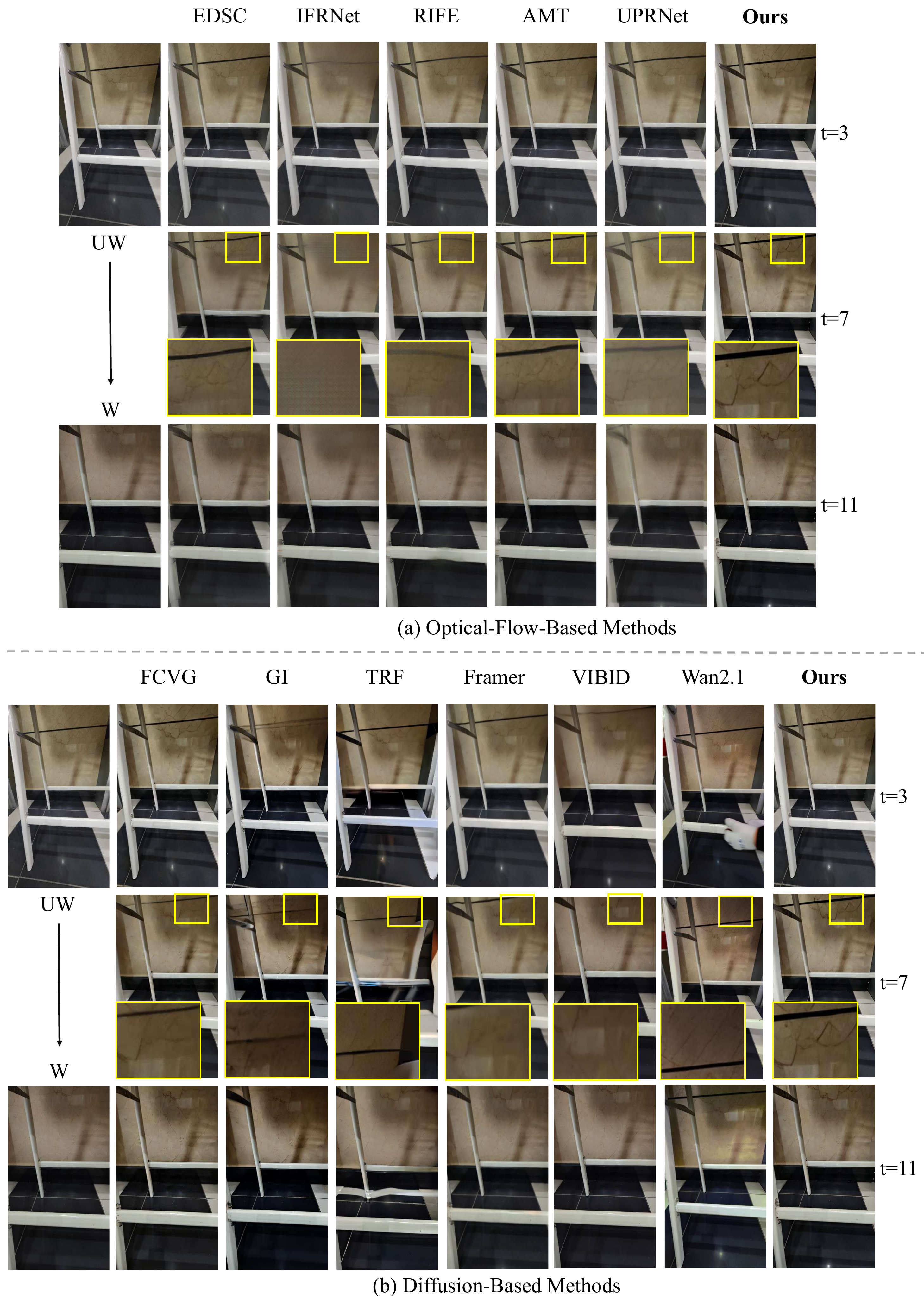}
\caption{
Visual comparison with state-of-the-art frame interpolation methods on a real-world scene.
(a) Optical-flow-based methods.
(b) Diffusion-based methods.
Our ZoomDiff generates more geometrically consistent zoom transitions with fewer artifacts (yellow boxes).
Please zoom in for better observation.
}
\label{fig:compare_real2}
\end{figure*}

We also present additional qualitative comparisons with 12 state-of-the-art frame interpolation methods, including 5 optical-flow-based ones (i.e., EDSC~\cite{EDSC}, IFRNet~\cite{IFRNet}, RIFE~\cite{RIFE}, AMT~\cite{AMT}, UPRNet~\cite{UPRNet} and 6 diffusion-based ones (i.e., FCVG~\cite{FCVG}, GI~\cite{GI}, TRF~\cite{TRF}, VIBID~\cite{vibid}, Framer~\cite{framer} and Wan2.1~\cite{wan}), on both synthetic and real-world datasets.
As shown in Fig.~\ref{fig:compare_syn1}, Fig.~\ref{fig:compare_syn2}, Fig.~\ref{fig:compare_real1}, and Fig.~\ref{fig:compare_real2}, existing methods often produce geometric distortions and texture artifacts during zoom transitions, while ZoomDiff generates more geometrically consistent results with sharper details.

\section{Sec.B More Ablation Results}
% \section{Sec.C}
In addition to the quantitative ablation results reported in the main paper, we provide more qualitative comparisons to analyze the effects of different design components in ZoomDiff.
\subsection{Effect of Dual-Image Utilization in Latent Space}
\begin{figure*}[t!]    
\centering    
\includegraphics[width=0.90\textwidth]{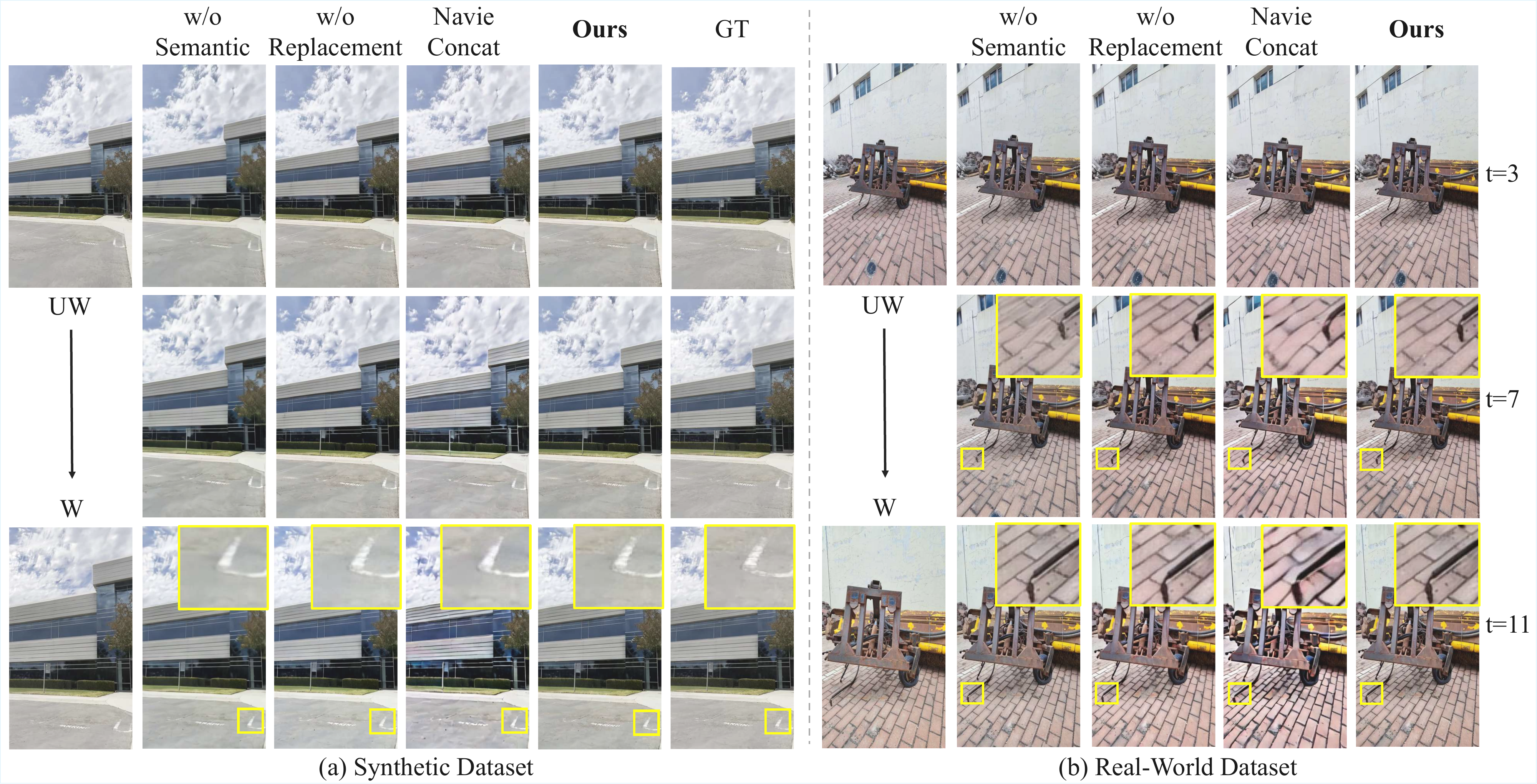}    
\caption{    
Visual effect of different strategies to inject dual images into the latent space.
Please zoom in for better observation.
}    
\label{fig:latent}
\end{figure*}

We provide qualitative comparisons of different latent-space utilization strategies in Fig.~\ref{fig:latent}. 
Compared with our full model, removing semantic guidance (i.e., w/o Semantic) results in less accurate structural details, demonstrating that high-level semantic information from the reference images is important for preserving scene content during denoising. 
Without step-wise latent replacement (i.e., w/o Replacement), the boundary frames gradually deviate from the input references, causing brightness inconsistency and accumulated artifacts along the zoom trajectory. 
The naive concatenation strategy (i.e., Naive Concat) introduces unstable conditioning across frames, leading to inferior visual quality and inconsistent transitions. 
By effectively combining semantic guidance and boundary latent anchoring, our method generates more coherent and visually consistent zoom transitions.

\subsection{Effect of Optimization Objectives}
\begin{figure*}[t!]    
\centering    
\includegraphics[width=0.90\textwidth]{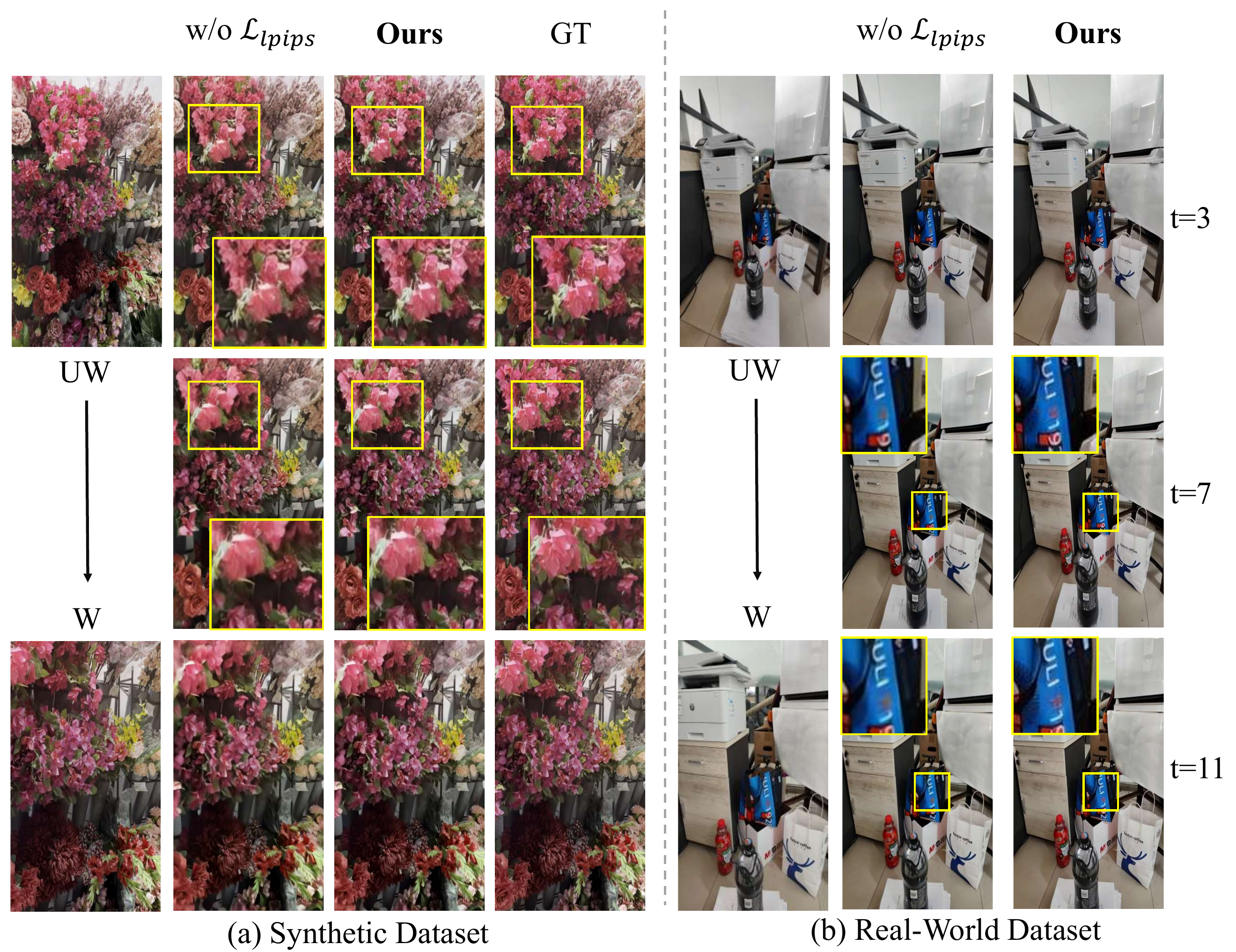}    
\caption{    
Visual effect of removing $\mathcal{L}_{lpips}$.
Removing $\mathcal{L}_{lpips}$ leads to degraded perceptual quality, causing distorted and blurry fine details.
With $\mathcal{L}_{lpips}$, the model recovers more realistic textures and fine-grained details.
}    
\label{fig:lpips}
\end{figure*}

\begin{figure*}[t!]    
\centering    
\includegraphics[width=0.95\textwidth]{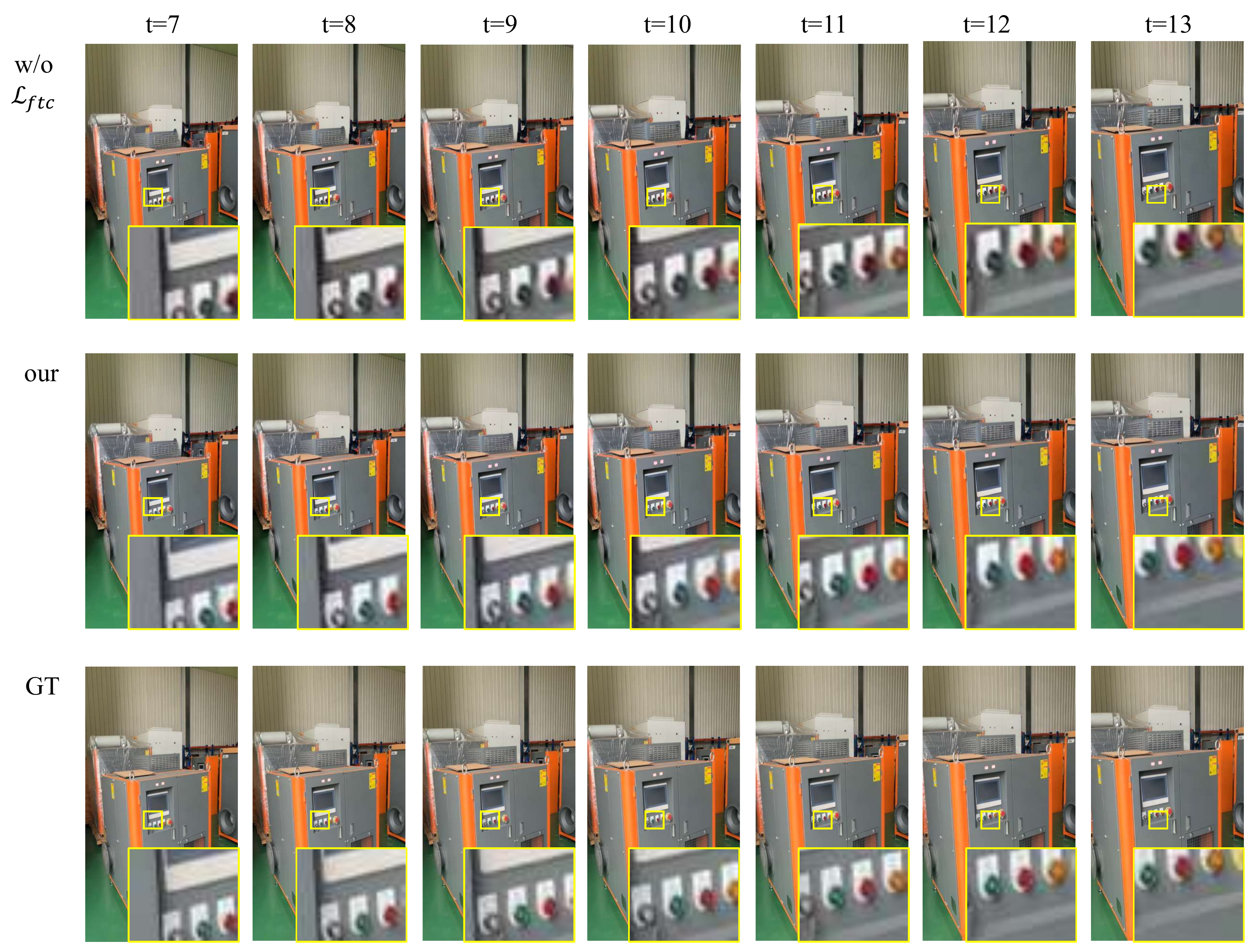}    
\caption{ 
Visual effect of removing $\mathcal{L}_{ftc}$.
Removing $\mathcal{L}_{ftc}$ reduces temporal consistency across consecutive frames, leading to gradual color and brightness deviations over time.
As shown in the yellow boxes, the indicator lights become darker as $t$ increases, while our method better preserves a stable appearance and smooth temporal transitions.
}    
\label{fig:ftc}
\end{figure*}

We provide additional qualitative comparisons of different loss configurations to further illustrate the effects of $\mathcal{L}_{lpips}$ and $\mathcal{L}_{ftc}$.
The visual results in Fig.~\ref{fig:lpips} and Fig.~\ref{fig:ftc} highlight the complementary roles of perceptual and temporal supervision.
As shown in Fig.~\ref{fig:lpips}, removing $\mathcal{L}_{lpips}$ leads to degraded perceptual quality, with distorted and blurry fine details in the generated frames.
Introducing $\mathcal{L}_{lpips}$ helps recover more realistic textures and preserve fine-grained details.
As shown in Fig.~\ref{fig:ftc}, removing the proposed flow-guided temporal consistency loss $\mathcal{L}_{ftc}$ causes color and brightness deviations across consecutive frames, which become more apparent as the zoom transition progresses.
With $\mathcal{L}_{ftc}$, the generated sequence better preserves consistent appearance and temporal coherence.
% We provide qualitative comparisons of different loss configurations in Fig.~\ref{fig:loss}.
% %
% Compared with using only the pixel-wise reconstruction loss, introducing $\mathcal{L}_{lpips}$ produces sharper and more realistic details, especially in texture-rich regions.
% %
% Further incorporating the proposed flow-guided temporal consistency loss $\mathcal{L}_{ftc}$ improves temporal coherence and reduces flickering artifacts during the zoom transition.
% %
% These visual results demonstrate the complementary effects of perceptual and temporal supervision.
% \begin{figure*}[t!]    
% \centering    
% \includegraphics[width=0.99\textwidth]{Figures/loss.pdf}    
% \caption{    
% Visual effect of different optimization objectives.
% Removing both $\mathcal{L}_{lpips}$ and $\mathcal{L}_{ftc}$ results in over-smoothed images with limited perceptual quality.
% Introducing $\mathcal{L}_{lpips}$ recovers more fine-grained details, while the proposed $\mathcal{L}_{ftc}$ further suppresses temporal flickering and produces smoother video transitions.
% }    
% \label{fig:loss}
% \end{figure*}
% \begin{figure*}[t!]    
% \centering    
% \includegraphics[width=0.90\columnwidth]{Figures/lpips.pdf}    
% \caption{    
% Visual effect of removing $\mathcal{L}_{lpips}$.
% % 
% Removing $\mathcal{L}_{lpips}$ leads to degraded perceptual quality, causing distorted and blurry fine details.
% % 
% With $\mathcal{L}_{lpips}$, the model recovers more realistic textures and fine-grained details.
% }    
% \label{fig:lpips}
% \end{figure*}
\section{Sec.C More Challenging Cases}
\begin{figure*}[t!]    
\centering    
\includegraphics[width=0.80\textwidth]{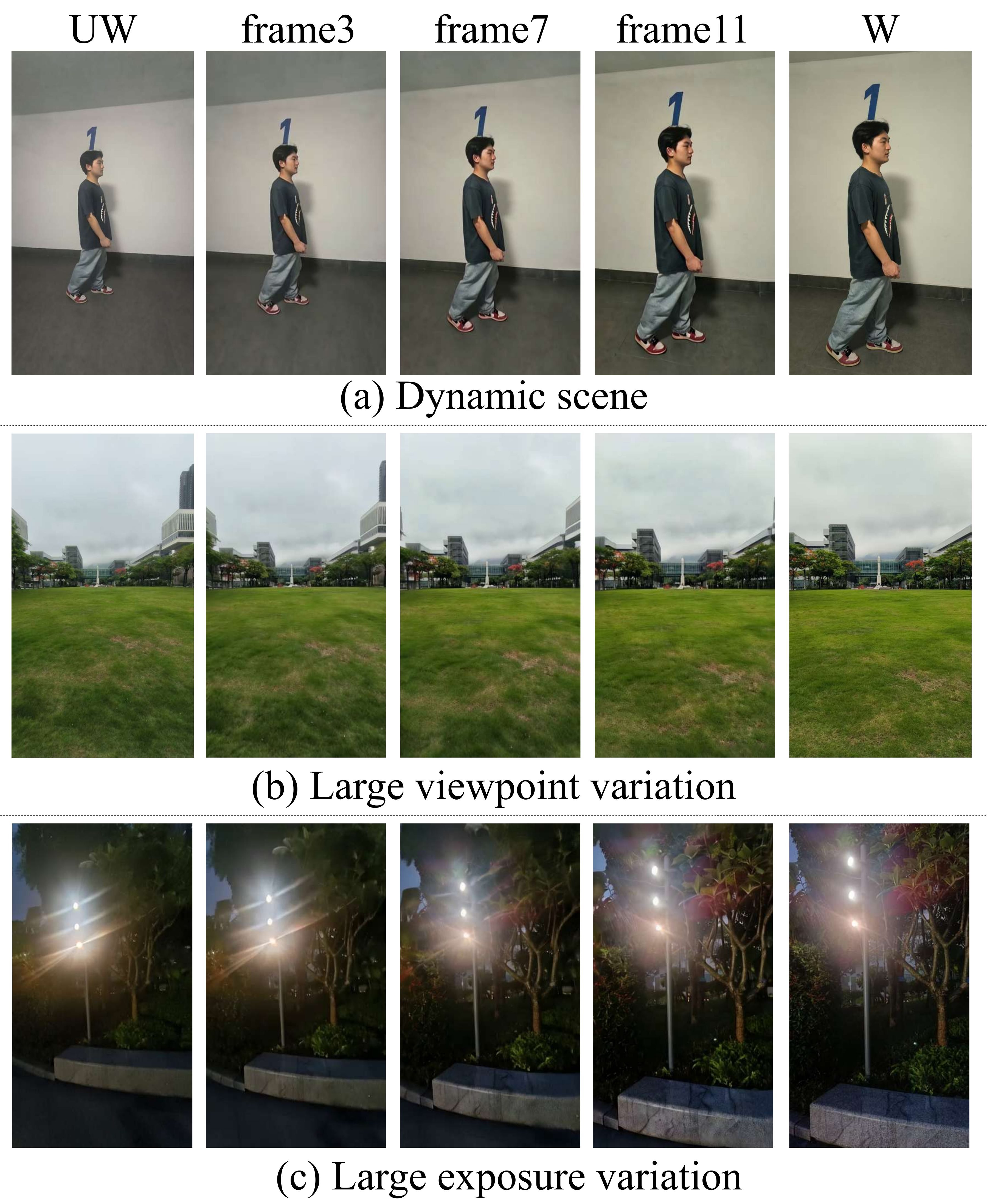}   
\caption{    
Evaluation on challenging scenes.
Please zoom in for better observation.
}    
\label{fig:challenging}
\end{figure*}

% \section{Sec.D}
% We present additional challenging cases in real-world scenarios, including human motion, large FOV changes, and severe exposure variations. 
% Fig.~\ref{fig:challenging}(a)(i) shows a scene with human motion, while Fig.~\ref{fig:challenging}(a)(ii-iii) demonstrate large FOV differences and severe exposure changes. 
% Our method maintains smooth transitions and strong temporal and geometric consistency under these challenging conditions.
% 
We present additional challenging cases in real-world scenarios, including human motion, large FOV changes, and severe exposure variations.
Fig.~\ref{fig:challenging}(a)(i) shows a scene with dynamic human motion, where moving objects introduce additional temporal ambiguity during the zoom transition.
Fig.~\ref{fig:challenging}(a)(ii) demonstrates a large FOV change between dual cameras, requiring the model to handle substantial viewpoint variation and content differences.
Fig.~\ref{fig:challenging}(a)(iii) presents a case with severe exposure variation, where significant illumination discrepancies between the two cameras make consistent appearance preservation challenging.
Despite these challenging conditions, our method generates smooth transitions while maintaining strong temporal consistency and geometric coherence.
% \begin{figure*}[t!]    
% % \centering    
% \includegraphics[width=0.99\columnwidth]{Figures/challenging_case.pdf}   
% \caption{    
% Visual effect of different strategies to inject dual images into the latent space.
% % 
% Please zoom in for better observation.
% }    
% \label{fig:challenging}
% \end{figure*}
% % 